\documentclass[letterpaper, 10 pt, conference]{ieeeconf}  

\IEEEoverridecommandlockouts                              

\usepackage{graphicx} 
\usepackage{epsfig} 
\usepackage{mathptmx} 
\usepackage{times} 
\usepackage{amsmath} 
\usepackage{amssymb}  
\usepackage{cite}

\title{\LARGE \bf
A Thermoplastic Elastomer Belt Based Robotic Gripper*
}

\author{Xingwen Zheng$^{1}$, Ningzhe Hou$^{2}$, Pascal Johannes Dani$\ddot{e}$l Dinjens$^{3}$, Ruifeng Wang$^{2}$,  \\Chengyang Dong$^{2}$, and Guangming Xie$^{1,4,5}$
\thanks{*This work was supported in part by grants from the National Natural Science Foundation of China (NSFC, No. 91648120, 61633002, 51575005) and the Beijing Natural Science Foundation (No. 4192026).}
\thanks{$^{1}$Xingwen Zheng and Guangming Xie are with the State Key Laboratory of Turbulence and Complex Systems, Intelligent Biomimetic Design Lab, College of Engineering, Peking University, Beijing, 100871, China.
         {\tt\small (E-mail: \{zhengxingwen, xiegming\}@pku.edu.cn)}}%
         \thanks{$^{2}$Ningzhe Hou, Ruifeng Wang, and Chengyang Dong are with the Department of Electrical and Electronic Engineering, The University of Manchester, Manchester, M1 3BB, United Kingdom.
         {\tt\small (E-mail: \{ningzhe.hou, ruifeng.wang, chengyang.dong\}@student.manchester.ac.uk)}}
         \thanks{$^{3}$Pascal Johannes Dani$\ddot{e}$l Dinjens is with Fontys University of Applied Sciences, Eindhoven, 5612 AP, The Netherlands.
         {\tt\small (E-mail: p.dinjens@student.fontys.nl)}}
\thanks{$^{4}$Guangming Xie is with the Institute of Ocean Research, Peking University, Beijing, 100871, China.}%
\thanks{$^{5}$Guangming Xie is with the Peng Cheng Laboratory, Shenzhen, 518055, China}%
}

\begin{document}

\maketitle
\thispagestyle{empty}
\pagestyle{empty}

\begin{abstract}
Novel robotic grippers have captured increasing interests recently because of their abilities to adapt to varieties of circumstances and their powerful functionalities. Differing from traditional gripper with mechanical components-made fingers, novel robotic grippers are typically made of novel structures and materials, using a novel manufacturing process. In this paper, a novel robotic gripper with external frame and internal thermoplastic elastomer belt-made net is proposed. The gripper grasps objects using the friction between the net and objects. It has the ability of adaptive gripping through flexible contact surface. Stress simulation has been used to explore the regularity between the normal stress on the net and the deformation of the net. Experiments are conducted on a variety of objects to measure the force needed to reliably grip and hold the object. Test results show that the gripper can successfully grip objects with varying shape, dimensions, and textures. It is promising that the gripper can be used for grasping fragile objects in the industry or out in the field, and also grasping the marine organisms without hurting them.
\end{abstract}
\section{INTRODUCTION}
The traditional robotic grippers are designed to mimic the human hand. Gripping by means of fingers creating pressure points to hold the object. \cite{Lovchik1999The,crisman1996graspar,cabas2006optimized}. In recent years compliant grippers have become more realistic due to new materials and manufacturing processes sparking the interest of researchers. And the applications of robotic grippers in varieties of circumstances, especially in a harsh environment, have captured increasing interests.

Specifically, Kevin \textit{et al.} have developed a soft robotic gripper for sampling fragile species on deep reefs. The robotic gripper consists of bellow-style soft actuators. It has been used for collecting soft coral at a depth of 100 m in the sea \cite{galloway2016soft}. Zhi \textit{et al.} have manufactured a polyhedron-like gripper which can be actuated for rotating and self-folding. The gripper has been used to noninvasively enclose marine organisms in the sea at a depth of 700 m \cite{teoh2018rotary}. Shuguang \textit{et al.} have designed a magic ball inspired robotic gripper using vacuum-driven origami technology. The gripper is able to lift a variety of food and bottles with different weights, shapes, and sizes by means of skin friction and vacuum adsorption \cite{li2019vacuum}. Yufei \textit{et al.} have designed a four-finger air pressure actuated soft robotic gripper with variable effective lengths. The robotic gripper is able to grip objects with different shapes and sizes from 2mm to 170 mm \cite{hao2018soft}, even in amphibious environments \cite{hao2017modeling}. Pedro \textit{et al.} have developed a sea lamprey-like soft gripper that has a closed structure. Force of the gripper has been controlled and measured \cite{pedro2018closed}. Jeffrey \textit{et al.} have manufactured a soft gripper which is made of a flexible latex membrane sealed volume chamber. It grips objects using the friction between the membrane and the objects \cite{krahn2017soft}. Zhenishbek \textit{et al.} have designed an origami-inspired suction gripper with shape memory alloy actuators. It can be self-fold into three shapes for picking objects \cite{zhakypov2018origami}. Chih \textit{et al.} have used silicon rubber to manufacture a robotic gripper that can grip irregular objects \cite{liu2017optimal}. Yonghua \textit{et al.} have designed soft robotic grippers which use silicone rubber soft actuators and packs of particles to form stiffness-controllable fingers \cite{li2017passive,li2019soft,jiang2019variable}. The gripper is able to adjust its stiffness according to the external environments.

The above-mentioned research has greatly contributed to the development of novel robotic gripper research. Most of them have novel structures without finger-like actuators which have been typically in robotic gripper \cite{teoh2018rotary,li2019vacuum,pedro2018closed,krahn2017soft,zhakypov2018origami,liu2017optimal}. On the other hand, for novel gripper with fingers, their fingers are typically made of novel materials \cite{galloway2016soft,hao2018soft,hao2017modeling,li2017passive,li2019soft,jiang2019variable}. However, most of the existing soft robotic grippers are actuated by means of pneumatic or hydraulic systems. These systems come with complications such as a reservoir tank which takes space, or an active pump which leads to high energy use. Pistons used in pneumatic and hydraulic are generally quite expensive and bigger than electric motors. Soft gripper actuated with a single motor could be the choice for projects that have limited size, budget, or energy supply.
\begin{figure}[htb]
\centering
\includegraphics[width=\columnwidth]{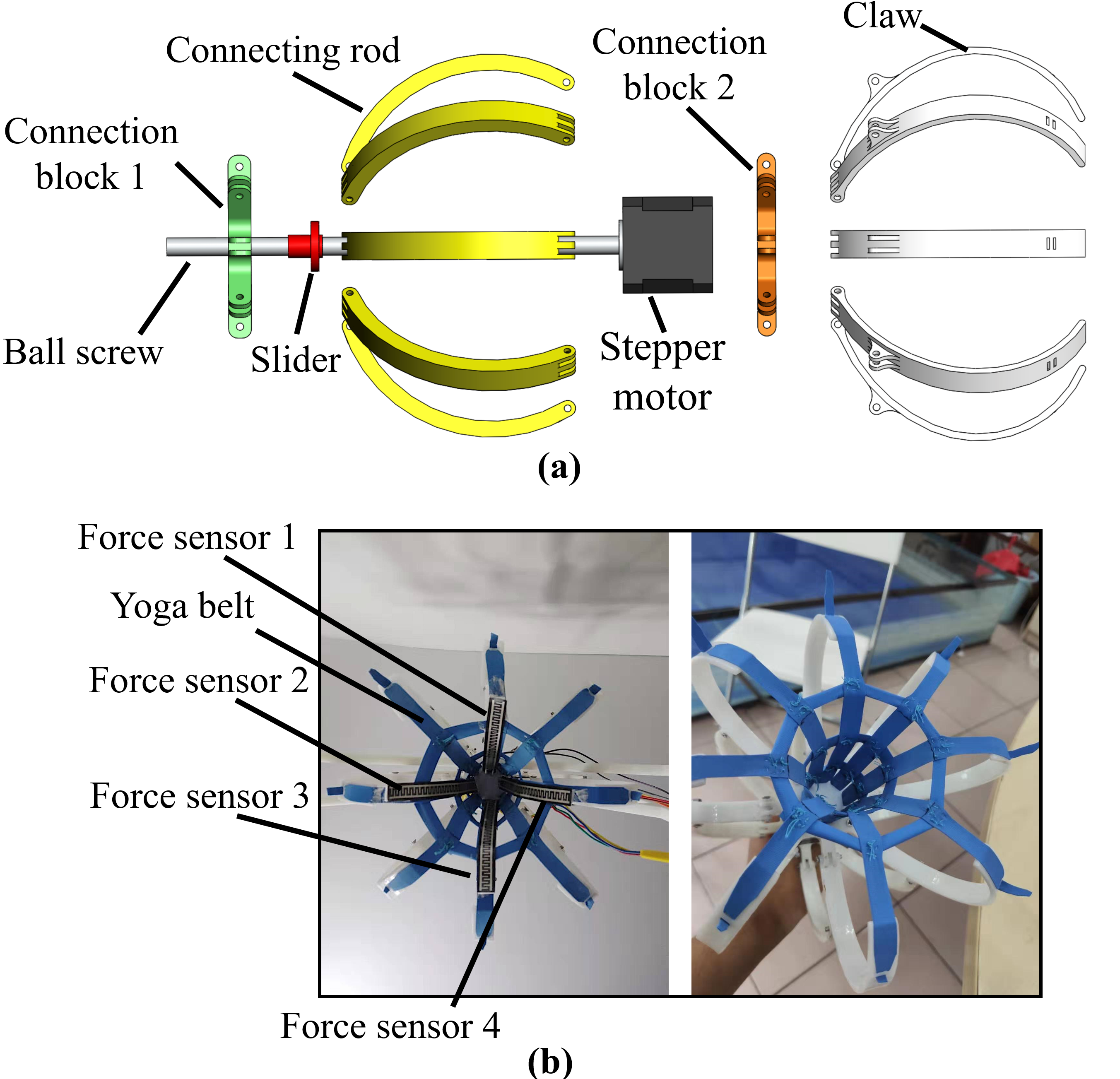}
\caption{Design of the gripper. (a) Configurations of the gripper. (b) The prototype of the gripper with the force sensors.}
\label{Design_of_the_gripper}
\end{figure}

Basing on the above analyses, in this paper, we design a novel robotic gripper consists of two external deformable frames and one internal soft thermoplastic elastomer belt-made net which is attached to one of the two frames. A motor is used for actuating the frames. The deformation of the frames caused the expansion and contraction of the thermoplastic elastomer belt-made net. The robotic gripper grip objects using the friction between the net and objects. Stress simulation has been firstly conducted for investigating the regularity between the normal stress on the net and the displacement of the net. Besides, force sensors are attached to the net for measuring the force needed for gripping objects, and the force data are analyzed in detail. In addition, the advantage and disadvantage of the gripper, the limitation of the experiments, and the potential applications of the gripper have been discussed in detail. It is promising to apply the designed gripper for grasping fragile objects and marine organisms without damage.

The remainder of this paper is organized as follows. Section II exhibits the design of the gripper. Section III describes the setup for force measurement experiment of the gripper. Section IV introduces the experimental results and analyses. Section V concludes this article and proposes future work.
\begin{figure*}[!htb]
\centering
\includegraphics[width=\linewidth]{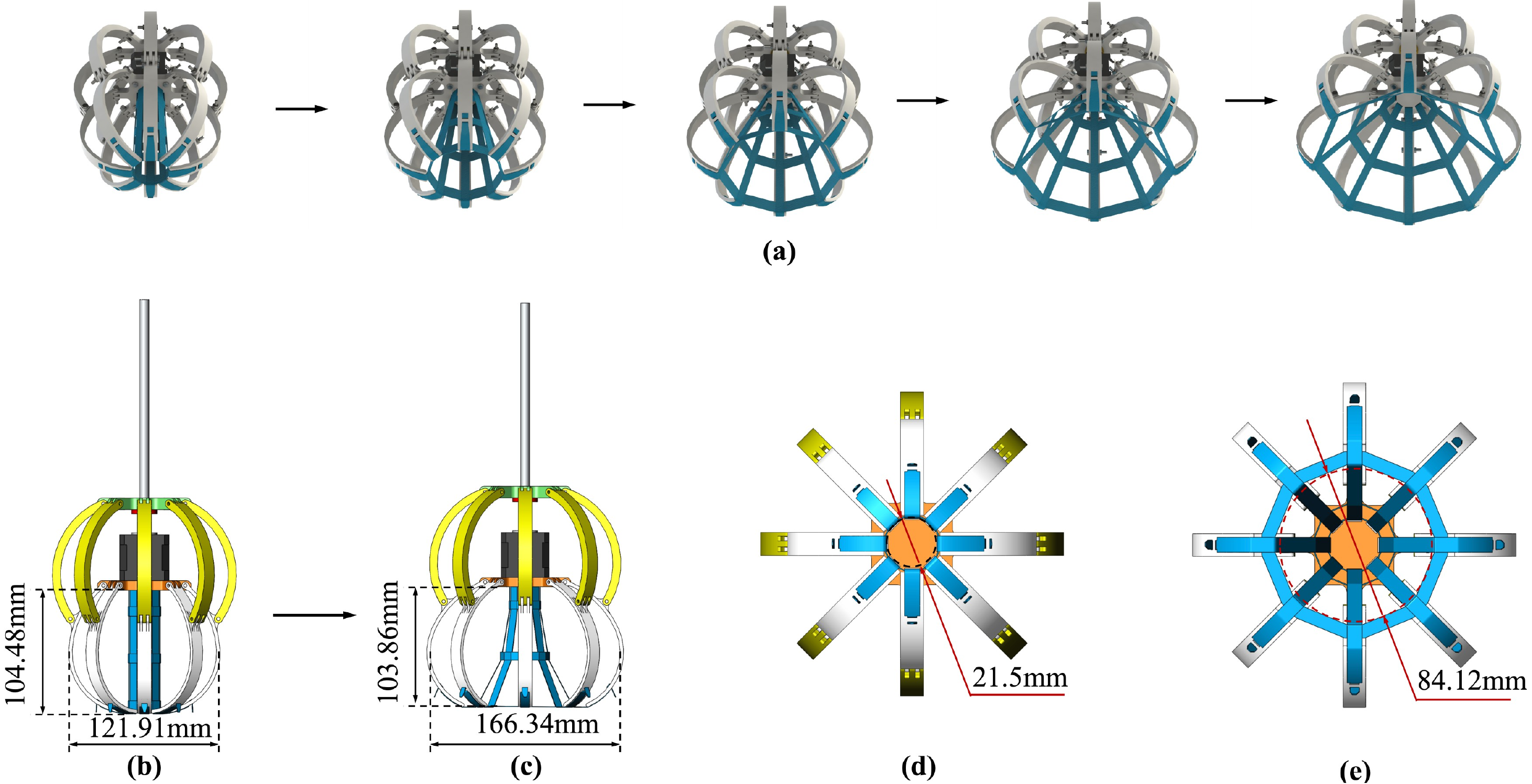}
\caption{Closed and open states of the gripper. (a) The process of opening the gripper. (b) Dimensions of the closed gripper. (c) Dimensions of the open gripper.(d) Dimensions of the closed thermoplastic elastomer belt-made net. (e) Dimensions of the open thermoplastic elastomer belt-made net.}
\label{Design_of_the_gripper2}
\end{figure*}
\begin{figure*}[!htb]
\centering
\includegraphics[width=\linewidth]{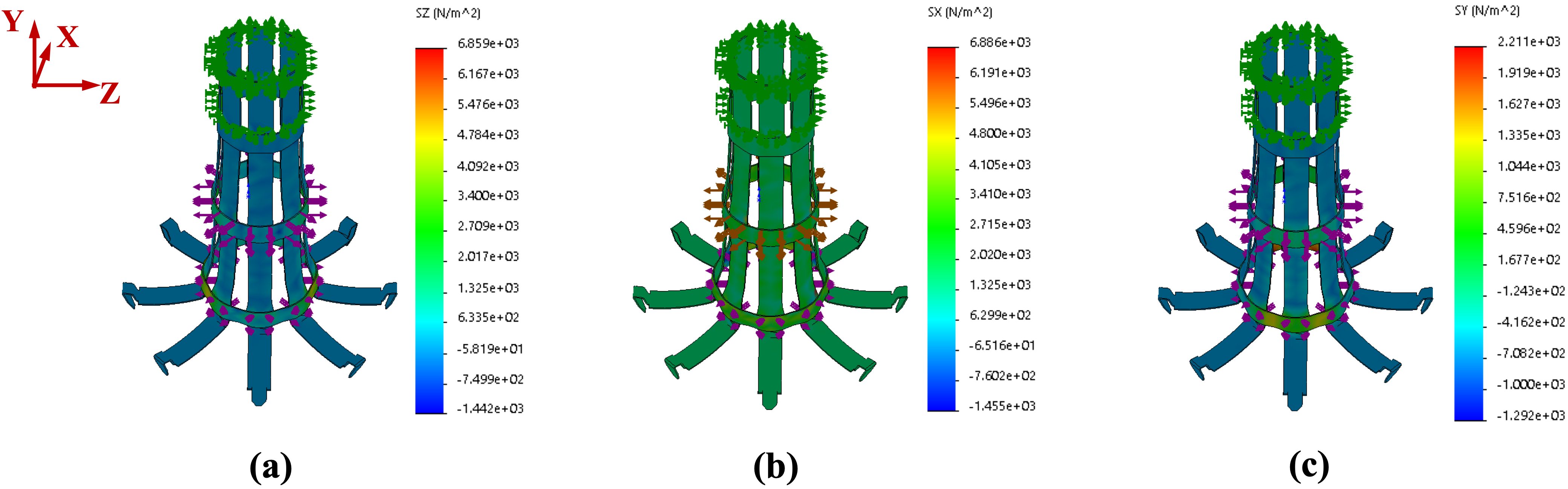}
\caption{Normal stress analysis for the thermoplastic elastomer net of the gripper. (a) X component of the normal stress on the net. (b) Y component of the normal stress on the net. (c) Z component of the normal stress on the net.}
\label{STRESS}
\end{figure*}
\begin{figure}[htb]
\centering
\includegraphics[width=0.8\columnwidth]{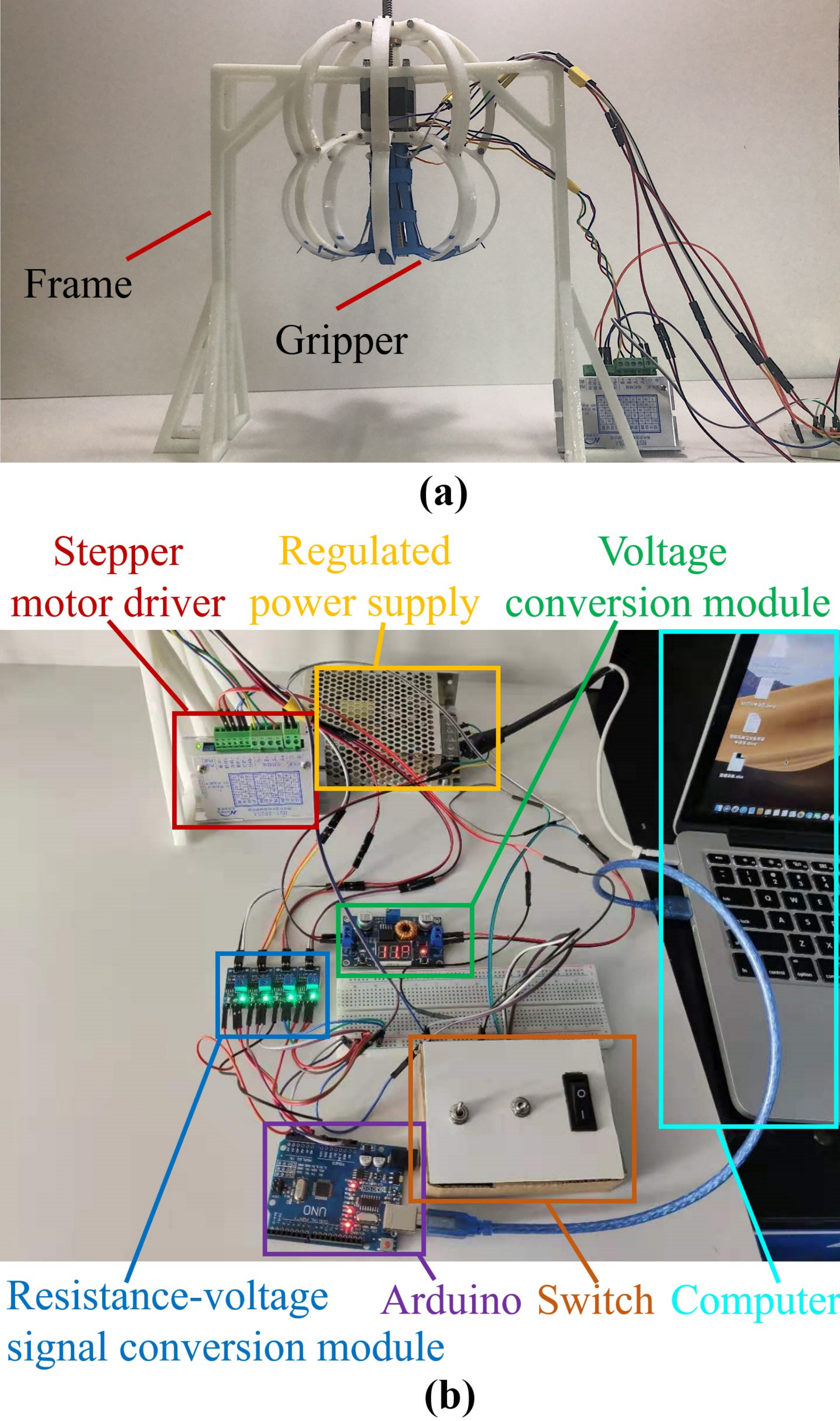}
\caption{Scene of test experiments of the gripper. (a) The gripper fixed on a frame. (b) Test circuit of the gripper.}
\label{Scene_of_test_experiments_of_the_gripper}
\end{figure}
\section{Design of the Gripper}
As shown in Figure~\ref{Design_of_the_gripper} (a), the gripper consists of a stepper motor with a ball screw and eight connecting rods which are separately connected with eight claws. The connecting rods and the claws are separately connected with central connection blocks 1 and 2. Connection block 1 is connected with a slider that is attached to the ball screw. Connection 2 is connected with the stepper motor. As shown in Figure~\ref{Design_of_the_gripper} (b), a thermoplastic elastomer belt-made reticulate elastic net is attached with the ends of the eight claws and the connection block 2. We have compared varieties of materials like nylon cord, cloth, silicone rubber mould, thermoplastic elastomer, etc. Finally, we have chosen thermoplastic elastomer belt because it has better elasticity than nylon cord and cloth, and it is easier to form a net than silicone rubber mould which needs careful operations during perfusion. Besides, four piezoresistive sensors are attached on the thermoplastic elastomer belt-made net, for measuring the touch force when the gripper is gripping an object.

Figure~\ref{Design_of_the_gripper2} (a) shows the process of opening the gripper. Through controlling the stepper motor, the slider is able to move along the ball screw, thus actuating the eight connecting rods. Motions of the connecting rods actuate the eight claws for opening or closing. When the claws open or close, the thermoplastic elastomer belt-made net will also stretch with different extents. More about the above-mentioned process can be found in the \textbf{supplementary video}. Figure~\ref{Design_of_the_gripper2} (b) and (c) show the dimensions of the gripper when it closes or opens to the maximum extent. Figure~\ref{Design_of_the_gripper2} (d) and (e) show the dimensions of the thermoplastic elastomer belt-made net. When the gripper closes to the maximum extent, it is able to grip an object whose size is bigger than 25 mm. While it can only grip object with a dimension that is less than 84.12 mm, when the gripper opens to the maximum extent. From the closed state to the open closed state, the slider moves about 9 mm along the ball screw. We will test the gripper for gripping varieties of objects in the following part. The connection rods, connection blocks, and claws of the gripper are made using SLA (Stereolithography) 3D Printer (iSLA550) from ZRapid Tech Company.
\section{Normal Stress Analysis for the Thermoplastic Elastomer-Made Net}
Normal Stress Analysis for the thermoplastic elastomer net of the soft gripper is conducted using SolidWorks Static Study. The objective of the simulation is to investigate the relationship between the displacement of the net corresponding to the volume of gripping object and the normal stress been applied to the object. As shown in Figure~\ref{STRESS}, the green arrows indicate the fix geometry and the purple and orange arrows indicate external force on the net. The force along the purple arrows is bigger than that along the orange arrows. It can be seen that, the normal stress increase with the displacement of the net. It can be inferred that, bigger force is needed for grasping objects with bigger volume.
\section{Experiment of Measuring the Grasping Force of the Gripper}
\subsection{Setup for the Experiment}
\begin{figure}[htb]
\centering
\includegraphics[width=\columnwidth]{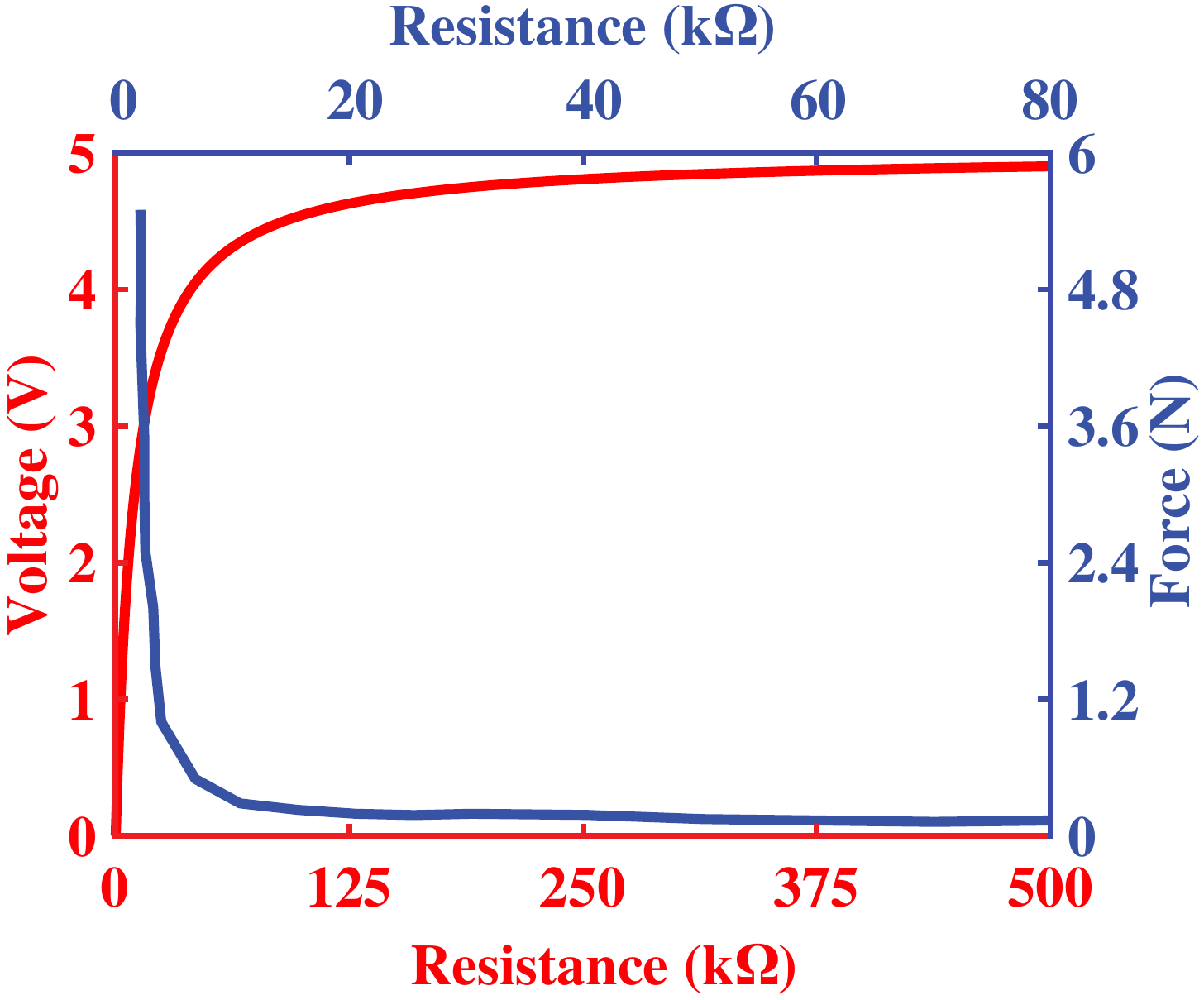}
\caption{Regularities among the residence of the piezoresistive sensor, output voltage, and the touch force.}
\label{Regularities}
\end{figure}
To test the touch force of the gripper when it is gripping objects, we have conducted multiple force measurement experiments. It is typical to fix the gripper to a robotic arm, then conducting experiments. Considering we have no robotic arms in the lab, we have fixed the gripper with a 3D-printed frame, as shown in Figure~\ref{Scene_of_test_experiments_of_the_gripper} (a). Such a design can suppress the measurement noise caused by the shake of the gripper. The force measured in the experiments refers to the normal force. In the experiments, the gripper has been controlled by sending position signal to the stepper motor. Once the objects have been completely held, we stopped the stepper motor. After force data recording, the stepper motor started again and the objects left the gripper.

Figure~\ref{Scene_of_test_experiments_of_the_gripper} (b) shows the test circuit of the gripper. It mainly includes a stepper motor driver, a regulated power supply device, a voltage conversion module, a resistance-voltage signal conversion module, an Arduino UNO, a switch, and a computer. The regulated power supply device is used for supplying electricity for the whole test circuit. The voltage conversion module is used to transfer the output voltage of the regulated power supply device from 12 V to 5V for the resistance-voltage signal conversion module and Arduino board. The force sensor we have used is a piezoresistive sensor which is a flexible strain gauge. When the external force is applied to the strain gauge, its resistance varies. A resistance-voltage signal conversion module has been chosen for transferring the resistance signal to voltage signal which has been monitored and saved in the computer. The regularities among the output voltage, the resistance signal, and the touch force has been measured and shown in Figure~\ref{Regularities}. The output voltage increases with the resistance. When there is no force applied on the sensor, the output voltage is 5 V. Basing on the regularities, the recorded output voltage can be transferred into the touch force.
\subsection{Experimental Results}
\begin{figure}[htb]
\centering
\includegraphics[width=\columnwidth]{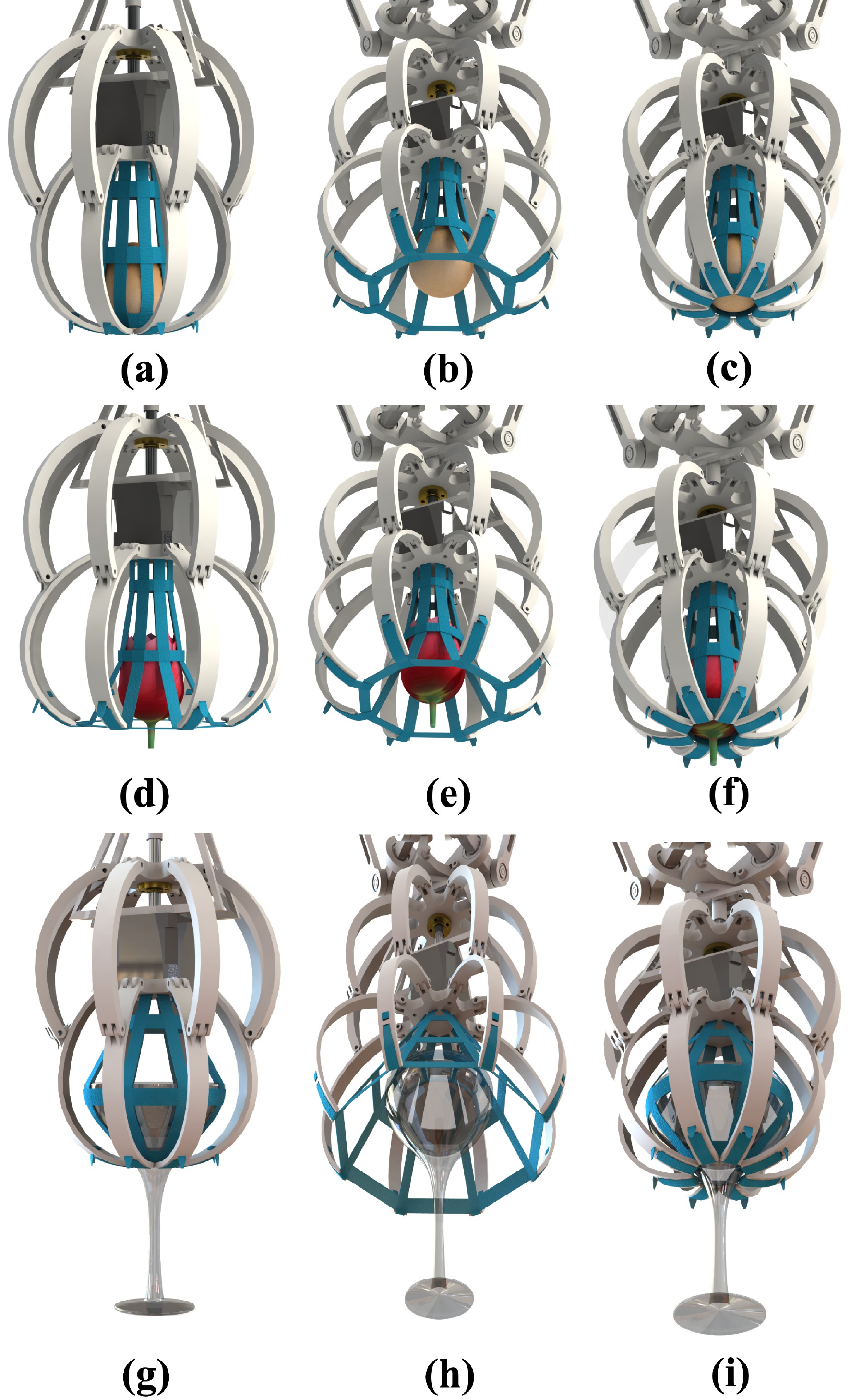}
\caption{Detailed processes of grasping egg, rose, and glass.}
\label{real}
\end{figure}

\begin{figure*}[htb]
\centering
\includegraphics[width=\linewidth]{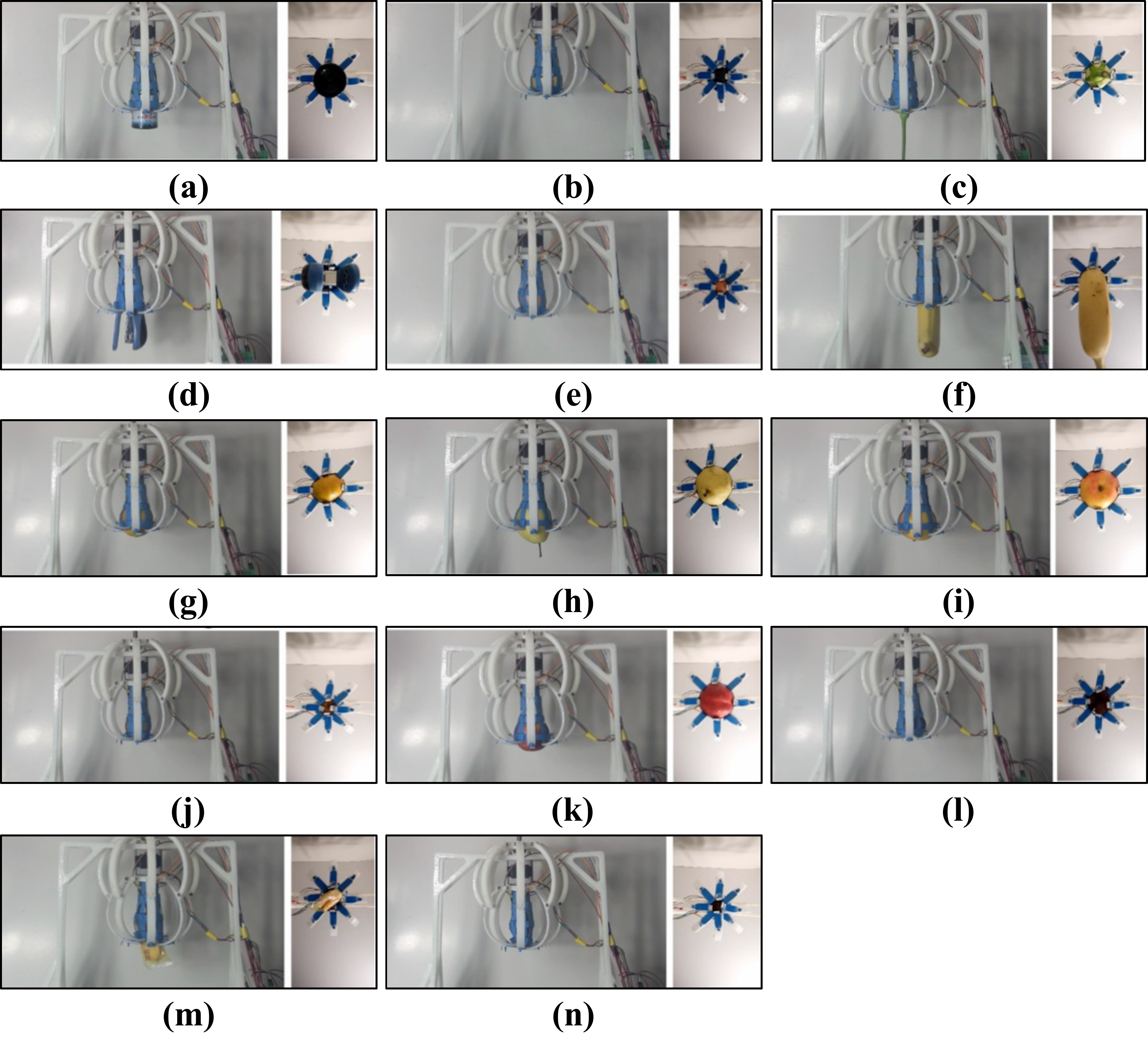}
\caption{Test of the gripper for different objects. (a) Bottle. (b) Tape. (c) Rose. (d) Stapler. (e) Egg. (f) Banana. (g) Orange. (h) Pear. (i) Apple. (j) Cherry tomato. (k) Peach. (l) Three grapes. (m) Biscuit. (n) Plum. For the whole process of gripping the above objects, please refer to the \textbf{supplementary video}.}
\label{Test_of_the_gripper_for_different_objects}
\end{figure*}

In the experiments, we have used the gripper to grip 20 species of objects with different shapes, dimensions, and textures. The objects include bottle, tape, rose, etc. In order to better exhibit the closing and opening of the gripper, Figure~\ref{real} shows detailed processes of grasping egg, rose, and glass using SolidWorks. Figure~\ref{Test_of_the_gripper_for_different_objects} has shown the real scenes of grasping most parts of the objects. More experimental scenes of gripping more objects can be found in the \textbf{supplementary video}.

\begin{figure*}[!htb]
\centering
\includegraphics[width=\linewidth]{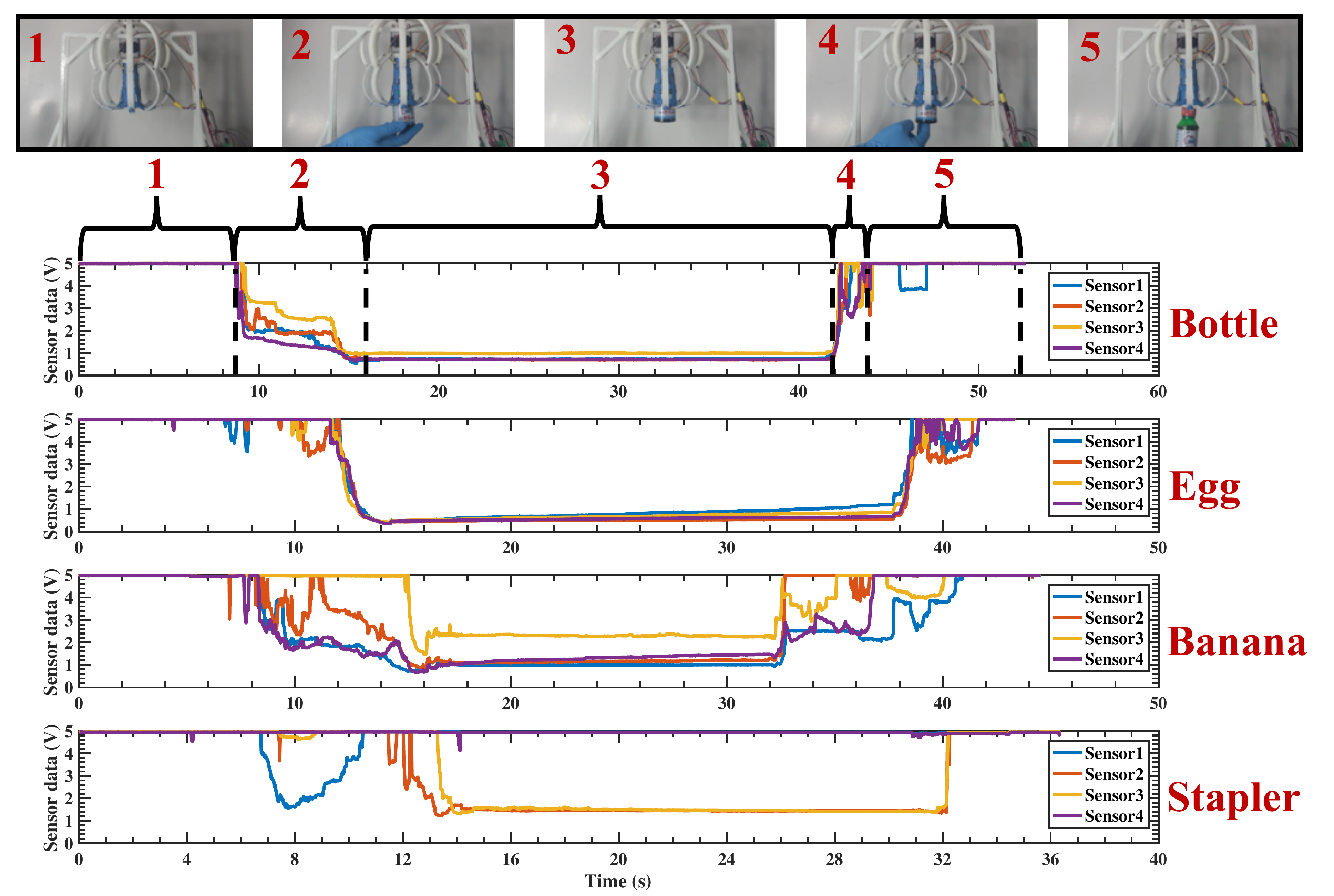}
\caption{The output voltage of the residence of the piezoresistive sensor with the time for different objects.}
\label{Output_voltage_time_different_objects}
\end{figure*}

In the experiments, we recorded the data of the force sensors for the whole process of gripping objects. Figure~\ref{Output_voltage_time_different_objects} shows the sensor data with the time for gripping bottle, egg, banana, and stapler. Taking the sensor data of gripping bottle for example, we have divided the data curve into five steps. At the first step, the sensor data remained at 5 V, which means that the gripper did not touch the bottle. Once the gripper touched the bottle and closes (Step 2), the sensor data decreased with time. Until the gripper held the bottle, the sensor data remained at certain values (Step 3). In the experiments, we measured the force with which the objects were just held without damage. At the fourth step (Step 4), the gripper open, and the objects leaved with increasing sensor data. Once the objects completely left the gripper, the sensor data remained at 5 V again (Step 5). For gripping banana and stapler, there existed significant differences among the four sensors in Step 3. That is because sensor 3 did not touch the banana sufficiently as the other three sensors, for the case of gripping banana. For sensors 1 and 4 in the case of gripping stapler, they did not touch the stapler, so their data remained at 5 V.

The above mentioned has demonstrated that it is able to by monitoring  the force data with the time. Besides, the analysis of force data with the time has also provided a guidance for real-time force feedback based loop control of the gripper in the future work.

\begin{figure}[htb]
\centering
\includegraphics[width=\linewidth]{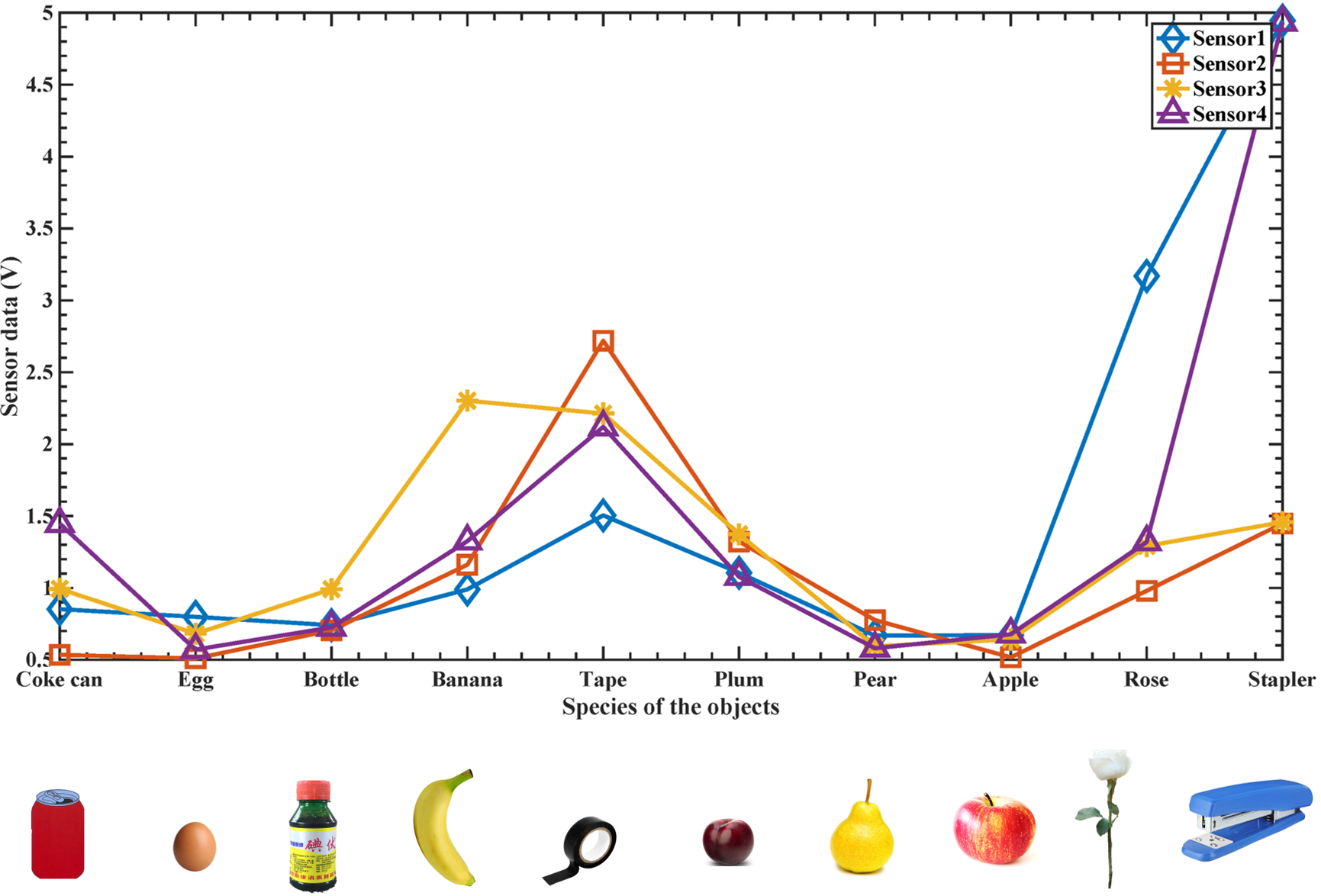}
\caption{Output voltage of the force sensor for gripping different objects.}
\label{Output_voltage_different_objects}
\end{figure}

\begin{figure}[htb]
\centering
\includegraphics[width=\linewidth]{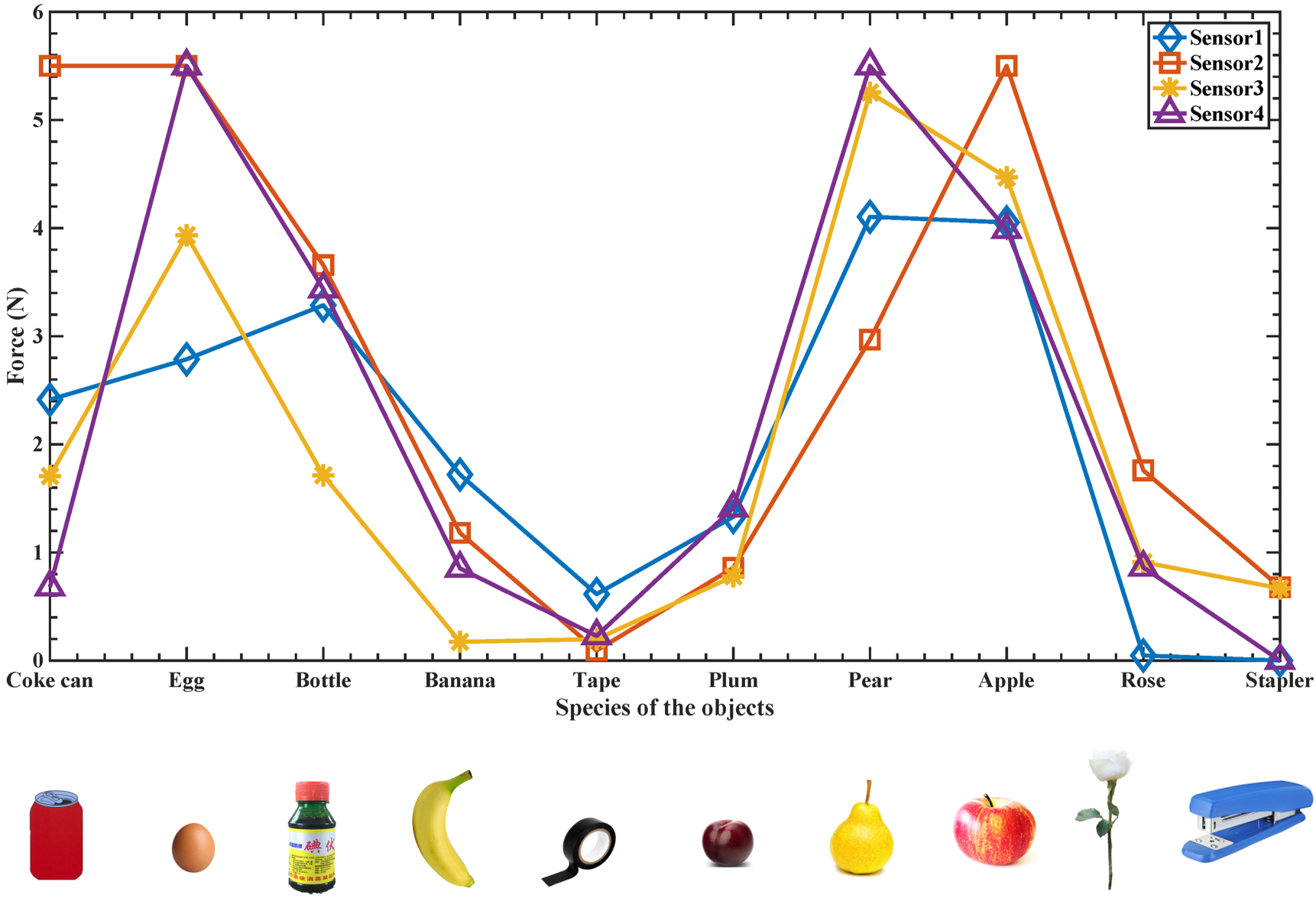}
\caption{The needed force for gripping different objects.}
\label{FORCE}
\end{figure}

We have also calculated the mean of the sensor data measured at step 3 for ten species of objects, as shown in Figure~\ref{Output_voltage_different_objects}. The pictures of the objects at the bottom of Figure~\ref{Output_voltage_different_objects} are shown on the same scale. Basing on Figure~\ref{Regularities}, we can obtain the force needed for gripping different species of objects, as shown in Figure~\ref{FORCE}. When objects were gripped by the gripper, their weights also had effects on extents of deformation of the piezoresistive sensor, thus affecting variations of the force data. Comparing with other objects, coke can, tape, plum, banana, and rose were lighter, so the forces needed were smaller. For stapler, because of its irregular shape, it has not touched sensors 1 and 4, so the force measured remained about 0. A more inspection revealed that forces measured by the four sensors have significant differences. This is because of the objects' irregular shapes which caused the different extents of touch to the sensors.

By analysing the mean of the force for gripping the objects, it provides a guidance for determining the threshold of the grasping force, thus determining the threshold of the signal of actuating the stepper motor, and finally avoiding the damage of the objects held by the gripper.

\section{Discussion}
\subsection{Advantages and Disadvantages of the Designed Gripper}
\subsubsection{Advantages}
Comparing with the existing grippers actuated by pneumatic drive technology or hydraulic transmission technology, the designed gripper is easier to be actuated only by controlling a stepper motor. We agree that the pneumatic-driven and hydraulic-driven technologies are inevitable and necessary for the grippers used in some industrial applications. However, it typically needs complex components like air pump and hydraulic cylinder, some of which have big dimensions. The complex components have usually resulted in the high cost existing in the operations. Besides, sometimes the unsteady pneumatic and hydraulic transmission have also affected the gripper. For the gripper designed in this paper, it is actuated by motor without the problems mentioned above. When it comes to industrial application in which the works are extremely complex and hard, high-power motor can be considered.
\subsubsection{Disadvantages}
Because of the shape of the thermoplastic elastomer net exhibits like a circular cone. It is difficult for it to grasp an object with a bottom which is bigger than its top. Besides, the size of the opening is influenced by the deformation degree of the thermoplastic elastomer. Besides, it can not be too big to ensure the enough grasping force and frictional force for holding the objects. As a result, range of the dimensions of the objects which can be grasped is limited.
\subsection{Limitation of the Experiments}
In the experiments, the gripper grasped the objects in the air. In the following work, we will investigate how to use the gripper for picking up objects lying on a table. Such a work is more challenging for such a gripper. Besides, the measured voltage contains a mix of the internal forces in the elastic belt and the external forces from grasping. It is also useful to investigate how to isolate them in the future. In addition, the force sensor used in this paper does not have a high accuracy, which should also be improved in the future.

As shown in Figure~\ref{Regularities}, the resistance of the sensor decreases when the force increases. However, the decrease of the resistance is negligible when the force is higher than 5 N. It means that the sensor can not reflect the force bigger than 5 N, which results in the measurement error in the experiments.
\subsection{Potential Applications of the Designed Gripper}
\begin{figure*}[htb]
\centering
\includegraphics[width=\linewidth]{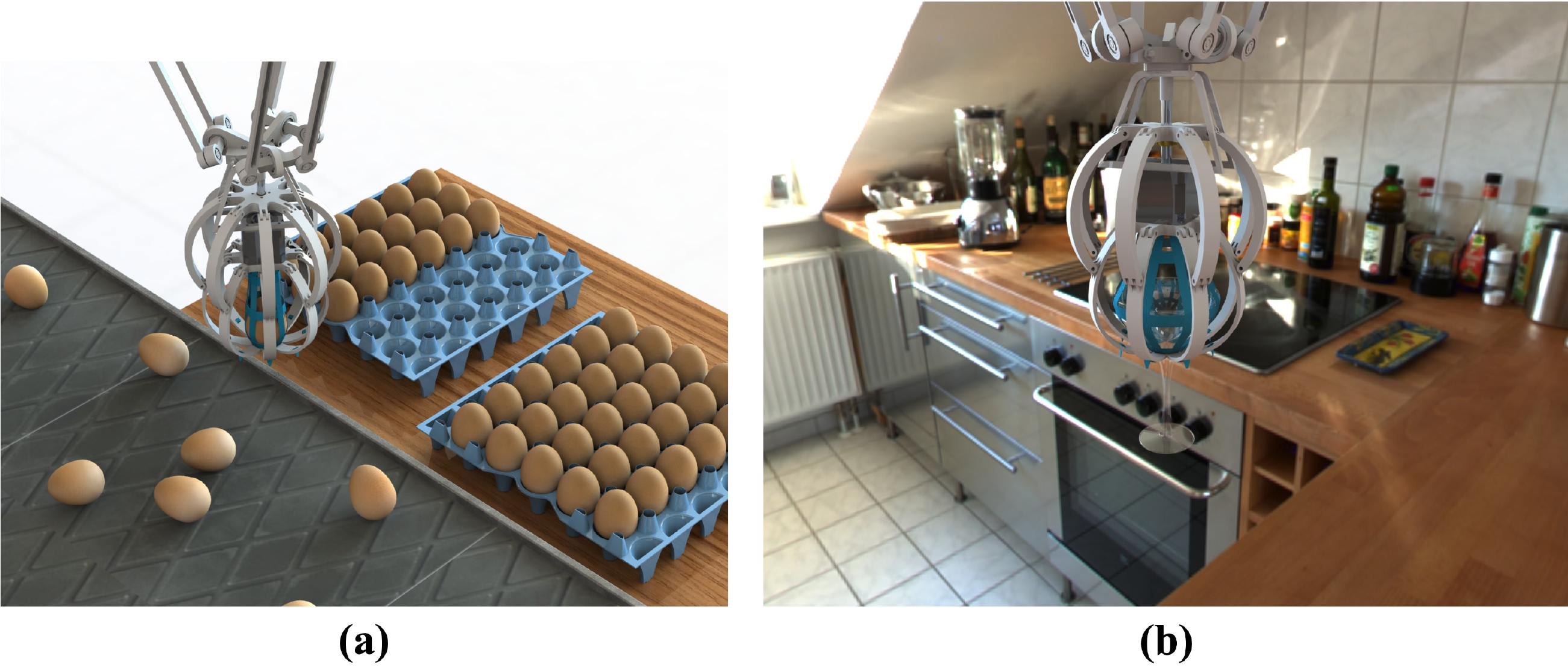}
\caption{Applications of the gripper in sorting eggs in the factory and glasses in the restaurant.}
\label{app}
\end{figure*}

Considering that the designed gripper has the ability of adaptive gripping through flexible contact surface of the thermoplastic elastomer belt, it can be used for grasping objects without damaging them. As shown in Figure~\ref{app}, the gripper can be attached to a robotic arm and then used in the line for sorting the eggs in the factory or used in the restaurant. Besides, the gripper can be also used for grasping marine organisms without damage. We am focusing on testing the gripper in real life scenarios.

\section{Conclusion and Future Work}
In this paper, a novel thermoplastic elastomer belt based robotic gripper has been designed. Besides, the circuit for actuating the gripper and data acquisition of the force sensor has been designed. Normal stress on the net with the deformation of the gripper has been explored. Besides, force measurement experiments have been conducted for determining the needed force of gripping varieties of objects with different shapes, dimensions, and textures, and results have been analyzed in detail. The experiments have demonstrated the efficiency of the gripper in grasping varieties of objects with different materials and dimensions.

In the future, we will modify the thermoplastic elastomer belt-made soft net and design the numbers and distributions of force sensors which will be attached on the soft net. Then we will conduct force feedback control of the robotic gripper, basing on the force regularity we will obtain. Besides, we will attempt to use the gripper for nondestructive sampling of marine organism.
\bibliographystyle{IEEEtran}
\bibliography{gripper_ref}
\end{document}